\documentclass{article}
\usepackage[utf8]{inputenc} % allow utf-8 input
\usepackage[T1]{fontenc}    % use 8-bit T1 fonts
\usepackage{hyperref}       % hyperlinks
\usepackage{url}            % simple URL typesetting
\usepackage{booktabs}       % professional-quality tables
\usepackage{amsfonts}       % blackboard math symbols
\usepackage{nicefrac}       % compact symbols for 1/2, etc.
\usepackage{microtype}      % microtypography

\usepackage{caption}
\usepackage{subcaption}

\usepackage{amsmath}
\usepackage{amssymb}
\usepackage{graphicx}
\usepackage[]{algorithm2e}

\title{Training products of expert capsules \\ with mixing by dynamic routing}
\author{%
  Michael Hauser \thanks{Michael Hauser has been supported by a postdoctoral fellowship at the Center for Autism Research in the Children's Hospital of Philadelphia. Any opinions, findings and conclusions or recommendations expressed in this publication are those of the author and do not necessarily reflect the views of the sponsoring agencies.} \\
 Children's Hospital of Philadelphia\\
  \texttt{mikebenh@gmail.com}
}
\date{}

\begin{document}

\maketitle

\begin{abstract}
This study develops an unsupervised learning algorithm for products of expert capsules with dynamic routing. Analogous to binary-valued neurons in Restricted Boltzmann Machines, the magnitude of a squashed capsule firing takes values between zero and one, representing the probability of the capsule being on. This analogy motivates the design of an energy function for capsule networks. In order to have an efficient sampling procedure where hidden layer nodes are not connected, the energy function is made consistent with dynamic routing in the sense of the probability of a capsule firing, and inference on the capsule network is computed with the dynamic routing between capsules procedure. In order to optimize the log-likelihood of the visible layer capsules, the gradient is found in terms of this energy function. The developed unsupervised learning algorithm is used to train a capsule network on standard vision datasets, and is able to generate realistic looking images from its learned distribution.
\end{abstract}

\section{Introduction}

Products of experts models~\cite{hinton2002training} were designed as an alternative to mixture models, where the sum over distributions is replaced by a product. Because there is a product, as opposed to a sum, if an individual expert votes a low probability, then the entire product  will have low probability. This is in contrast to a standard mixture model, where it only takes a single distribution to vote high probability for the data point to have high probability of existing, effectively ignoring the votes of all of the other distributions. A product of experts allows each expert to specialize on a single aspect of the data, and to disqualify a data point from existing only on this single aspect, whereas a sum of experts requires each distribution to model all aspects of the data, since within the sum of experts each distribution is essentially acting independently of the others.

%If the products of expert model has two distinct layers, where connections exist only across layers, not within layers, then the MCMC procedure for estimating the underlying distribution becomes very efficient. Additionally the contrastive divergence was shown to make an effective cost function to estimate the distribution.

Capsule networks~\cite{hinton2011transforming} with dynamic routing~\cite{sabour2017dynamic} depart in many significant ways from neural networks. Artificial neural networks originate from the McCulloch-Pitts model~\cite{mcculloch1943logical}, where a scalar neuron fires if the weighted sum of the neuron's scalar inputs reaches a threshold. For example, the outputs of the softmax and max-pool functions depend not only on the previous layer inputs, but also on the outputs from the other nodes at the current layer; besides these exceptions most standard activation functions, such as the sigmoid and ReLU, follow the McCulloch-Pitts design. 

As an alternative to scalar-valued neurons, capsules are vector valued, which has at least two advantages. First, because they are vector valued, much more complex routing mechanisms, such as routing by agreement~\cite{sabour2017dynamic}, can be used to pass data forward through the network, as opposed to a simple weighted sum of inputs. Second, vectors have a magnitude and an orientation, which allows capsules to fire if their magnitude reaches a threshold, while their orientation can be used to represent the instantiation parameters of the data.

A major difficulty with capsule networks is that because they are still in their infancy, far fewer algorithmic tools exist to train and develop them. For example capsule networks have been trained via backpropagation~\cite{rumelhart1985learning,hinton2011transforming,sabour2017dynamic} as well as expectation-maximization~\cite{hinton2018matrix}. For unsupervised learning they have been trained as an autoencoder~\cite{hinton2011transforming} as well as in a generative-adversarial setting~\cite{jaiswal2018capsulegan,goodfellow2014generative}, but these are built off of backpropagation, which itself was designed for neural networks without dynamic routing. In their supervised settings they have been paired with a fully-connected decoder~\cite{sabour2017dynamic,hinton2018matrix}, but this is to compel the network to learn invariant vector representations and help defend against adversarial attacks~\cite{goodfellow2014explaining}.

For these reasons we design a product of expert capsules in an analogous way to a product of expert neurons. From this, an efficient unsupervised learning procedure is developed to train this product of expert capsules in a bottom up fashion, using the dynamic routing procedure to mix the model distribution. With a given energy function compatible with dynamic routing, the contrastive divergence is minimized to learn the underlying density.

\section{Algorithm review}

This section will briefly review roduct of experts as well as dynamic routing between capsules.

\subsection{Product of experts learning}

As mentioned in the Introduction, product of experts models combine densities as products as opposed to sums~\cite{hinton2002training,welling2005exponential}. If we divide the experts into a visible layer and hidden layer of binary valued experts, the energy of this configuration is given as

\begin{equation}
-E\left(v,h\right) = 
\sum_{i,j} w_{ij} v_i,h_j +
\sum_{i} b_{i} v_i + 
\sum_{j} c_{j} h_j
\end{equation}

From this, the probability of a visible-hidden configuration is $ p\left(v,h\right)=\frac{1}{Z}e^{-E\left(v,h\right)} $, and marginalized over the hidden layer gives $ p\left(v\right) = \sum_h p\left(v,h\right)=\frac{1}{Z} \sum_h e^{-E\left(v,h\right)} $. Because there are no intra-layer connections, we can efficiently sample the hidden layer nodes as $p\left(h_j=1|v \right) = \nicefrac{1}{1+\exp{\left( - \sum_i w_{ij}v_j + c_j \right)} } $ and the visible layer nodes as $p\left(v_i=1|h \right) = \nicefrac{1}{1+\exp{\left( - \sum_j w_{ij}h_i + b_i \right)} } $. The gradient of the log-likelihood is given by

\begin{equation}
  \frac{\partial \log p}{\partial w_{ij}} \left(v\right) =
  p\left( h_j=1 | v \right) v_i - 
  \sum_{v}
  p\left(v\right)
  p\left( h_j=1 | v \right) v_i
\end{equation}

The first term on the right-hand side of this equation is generated by the data, while the second term is the expectation over the product of experts model itself. Truly estimating the second term with MCMC requires mixing the density to infinity. Instead one usually minimizes the contrastive divergence~\cite{hinton2002training}, which requires running the Markov chain once.

\subsection{Routing by agreement in capsule networks}

This section will briefly review capsule networks and dynamic routing~\cite{sabour2017dynamic}.

At layer $l$, given a collection $i=1,2,\dots,I$ of vector-valued capsules $ x^{(l)}_{i} $ and matrix-valued prediction maps $W^{(l)}_{ij}$, the pre-activation, vector-valued predicted capsule $j$ from capsule $i$ is $z^{(l+1)}_{j|i}=W^{(l)}_{ij}\cdot x^{(l)}_{i}$,
 where at layer $l+1$ we have the collection $j=1,2,\dots,J$ of capsules. We then take a weighted average of all of the predictions to yield the final, pre-activation, vector-valued capsule $j$ at layer $l+1$, i.e.
\begin{equation}
z^{(l+1)}_{j}=\sum_{i}c^{(l)}_{ij} W^{(l)}_{ij}\cdot x^{(l)}_{i}
\end{equation}
where the scalar-valued $c^{(l)}_{ij}$'s are determined by routing by agreement and each lower layer capsule can only make a finite amount of predictions, i.e. $\sum_{i}c^{(l)}_{ij}=1$.

Routing by agreement is a procedure that iteratively re-weighs each of the individual predictions to yield a final, collective prediction. If an individual prediction $z^{(l+1)}_{j|i}$ agrees well with the collective prediction $z^{(l+1)}_{j}$, then we would like to increase the routing weight $c^{(l)}_{ij}$ connecting these capsules. Agreement has been measured as both the inner product~\cite{sabour2017dynamic}, as well as the cosine distance~\cite{hinton2018matrix}, between the individual predictions and the squashed collective prediction. In this paper we use the cosine distance. The updated weights are then renormlized $\sum_{i}c^{(l)}_{ij}=1$, and the process is repeated. (If the cosine distance is used, then squashing can happen outside this loop since the cosine distance takes the inner product of unit vectors, so the magnitude isn't needed.)

The squashing function~\cite{sabour2017dynamic} we use is defined as follows:

\begin{equation}
    x^{(l+1)}_{j} = 
    \textnormal{squash}\left(z^{(l+1)}_{j} \right)=
    \frac{\Vert z^{(l+1)}_{j}\Vert^2}{1+\Vert z^{(l+1)}_{j}\Vert^2} 
    \frac{z^{(l+1)}_{j}}{\Vert z^{(l+1)}_{j}\Vert} 
\end{equation}

The intention of the squashing function is to scale the magnitude of $z^{(l+1)}_{j}$ between $0$ and $1$ while keeping the orientation the same. We can then interpret capsule $j$ is on if the magnitude of the squashing function is close to $1$, and $j$ is off if the magnitude is close to $0$:
\begin{equation}
    P\left(j=\textnormal{on}|x^{(l)}_1,\dots,x^{(l)}_{I}\right) = 
    \Vert \textnormal{squash}\left(z^{(l+1)}_{j} \right) \Vert
\end{equation}

% Additionally, since the orientation is decoupled from the magnitude, a single capsule can fire over the entire range of input orientations, as compared to a neuron which can only fire under its one specific oriented input.

Later in the paper we need the inverse map of the squash map, which we call unsquash:

\begin{equation}
    z^{(l+1)}_{j} = 
    \textnormal{unsquash}\left(x^{(l+1)}_{j} \right)=
    \sqrt{\frac{\Vert x^{(l+1)}_{j}\Vert}{1-\Vert x^{(l+1)}_{j}\Vert} }
    \frac{x^{(l+1)}_{j}}{\Vert x^{(l+1)}_{j}\Vert} 
\end{equation}{}

This is just the pre-image of the squashed vector so $ \left(\textnormal{unsquash} \circ \textnormal{squash}\right) \left(z\right) = \textnormal{identity}_{\mathbb{R}^d}\left(z\right) = z$, in a similar way as the logit map is the inverse map of the sigmoid.

\section{Capsule networks as product of experts}

This section develops the product of expert capsules model, where a capsule being either on or off, measured by the magnitude of the capsule vector after squashing, is analogous to the binary action of a neuron firing in a Restricted Boltzmann Machine.

We denote an individual capsule by $x^{(l)}_i$ for $i=1,2,\dots,I$. We also write $x^{(l)}=\left(x^{(l)}_1,x^{(l)}_2,\dots,x^{(l)}_I \right)$ to be the collection of all capsules on that layer. Similarly $\Vert x^{(l+1)}_j \Vert$ is the norm of capsule $j$ for $j=1,2,\dots,J$, and $\Vert x^{(l+1)} \Vert = \left(\Vert x^{(l+1)}_1 \Vert,\Vert x^{(l+1)}_2 \Vert,\dots,\Vert x^{(l+1)}_J \Vert \right)$.

\subsection{Energy function and conditional probabilities}

Before beginning we would like to make a subtle, yet important point. We note that if we want the magnitude of the squashed capsule to play an analogous role to the probability of a binary neuron firing, $P\left(\Vert x^{(l+1)}_j \Vert=1  |x^{(l)}\right)=\Vert \textnormal{squash}\left( z^{(l+1)}_j \right) \Vert$, for $z^{(l+1)}_j=\sum_i c^{(l)}_{ij}W^{(l)}_{ij}\cdot x^{(l)}_i$, the squashing part of the squashing function can be rewritten as a sigmoid activation, $ \frac{\Vert z^{(l+1)}_j \Vert^2}{1 + \Vert z^{(l+1)}_j \Vert^2} = \sigma\left( \log \Vert z^{(l+1)}_j \Vert^2 \right)$. Intuitively, $\Vert z^{(l+1)}_j \Vert \geq 0 $, whereas the sigmoid function should take arguments over all of $\mathbb{R}$, so taking the logarithm of the norm maps positive numbers to all of $\mathbb{R}$.

Define the energy across layers as follows:

\begin{equation}
\label{eqn:product-of-capsules-energy}
    E\left( x^{(l)},\Vert x^{(l+1)} \Vert \right) = 
    - \sum_j
    \log \left( \Vert \sum_i c^{(l)}_{ij}W^{(l)}_{ij}\cdot x^{(l)}_i \Vert^2 \right)
    \Vert x^{(l+1)}_j \Vert
\end{equation}
% \begin{multline}
%     \label{eqn:product-of-capsules-energy}
%     E\left( x^{(l)},x^{(l+1)} \right) = 
%     E_1\left( \hat{x}^{(l)}, \Vert x^{(l+1)} \Vert \right) +
%     E_2\left( \hat{x}^{(l+1)}, \Vert x^{(l)} \Vert \right) =
%     \\
%     \sum_j
%     \log \left( \Vert \sum_i c^{(l)}_{ij}W^{(l)}_{ij}\cdot \hat{x}^{(l)}_i \Vert^2 \right)
%     \Vert x^{(l+1)}_j \Vert
%     + 
%     \sum_i
%     \log \left( \Vert \sum_j c^{(l)}_{ij}W^{(l)T}_{ij}\cdot \hat{x}^{(l+1)}_j \Vert^2 \right)
%     \Vert x^{(l)}_i \Vert    
% \end{multline}
% This can be decomposed as follows:
The probability of a certain configuration $\left( x^{(l)},\Vert x^{(l+1)} \Vert \right)$ is then defined
\begin{equation}
    P\left(x^{(l)},\Vert x^{(l+1)} \Vert \right) = 
    \frac{1}{Z} \exp{\left( -
    E\left( x^{(l)},\Vert x^{(l+1)} \Vert \right)
    \right)}
\end{equation}
% \begin{multline}
%     P\left(\hat{x}^{(l)},\Vert x^{(l)} \Vert,\hat{x}^{(l+1)},\Vert x^{(l+1)} \Vert \right) = 
%     \frac{1}{Z} \exp{\left( 
%     E\left( x^{(l)},x^{(l+1)} \right)
%     \right)} = \\
%     \frac{1}{Z_1} \exp{\left( 
%     E_1\left( \hat{x}^{(l)},\Vert x^{(l+1)} \Vert \right)
%     \right)}
%     \frac{1}{Z_2} \exp{\left( 
%     E_2\left( \hat{x}^{(l+1)}, \Vert x^{(l)} \Vert \right)
%     \right)}
% \end{multline}
where $Z$ is the normalizing partition function. %Because we decoupled the dual roles of the capsules $x^{(l)}_i$ and $x^{(l+1)}_j$ into the pairs $\left(\hat{x}^{(l)}_i,\Vert x^{(l)}_i \Vert \right)$ and $\left(\hat{x}^{(l+1)}_j,\Vert x^{(l+1)}_j \Vert \right)$, we can marginalize over $ \left( \hat{x}^{(l+1)}, \Vert x^{(l)} \Vert \right) $. 
This is a product of expert capsules:

\begin{equation}
\label{eqn:capsules-prob-full}
    P\left(x^{(l)},\Vert x^{(l+1)} \Vert \right) =
    \frac{1}{Z}
    \Pi_j \left[
    \exp{\left(
    \log \left( 
    \Vert
    \sum_i c^{(l)}_{ij}W^{(l)}_{ij}\cdot x^{(l)}_i
    \Vert^2
    \right) \Vert x^{(l+1)}_j \Vert
    \right)}
    \right]
\end{equation}

We marginalize this distribution over the layer $l+1$ capsules firing:

\begin{equation}
\label{eqn:capsules-prob-marginalized}
    P\left(x^{(l)} \right) =
    \frac{1}{Z}
    \Pi_j \left[
    1 +
    \exp{\left(
    \log \left( 
    \Vert
    \sum_i c^{(l)}_{ij}W^{(l)}_{ij}\cdot x^{(l)}_i
    \Vert^2
    \right) 
    \right)}
    \right]
\end{equation}

In order to have an efficient means of sampling the hidden layer, it is necessary that at layer $l+1$ the nodes representing the capsules firing are not connected, so that the total probability is equal to the product of individual capsule probabilities, and so using equations~\ref{eqn:capsules-prob-full} and~\ref{eqn:capsules-prob-marginalized} we have:

\begin{equation}
\label{eqn:capsules-not-connected}
    P\left(\Vert x^{(l+1)} \Vert | x^{(l)}\right) =
    % \frac{P\left(\hat{x}^{(l)} , \Vert x^{(l+1)} \Vert\right)}{P\left(\hat{x}^{(l)}\right)} =
    \Pi_j
    P\left(\Vert x^{(l+1)}_j \Vert | x^{(l)}\right)
\end{equation}

From here it is straightforward to show that the conditional probability of capsule $j$ firing is equal to the magnitude of the squashed capsule:

\begin{equation}
\label{eqn:prob-consistent-with-squash}
P\left(\Vert x^{(l+1)}_j \Vert = 1 | x^{(l)}\right) =
\frac{\Vert \sum_i c^{(l)}_{ij}W^{(l)}_{ij}\cdot x^{(l)}_i \Vert^2 }{1+\Vert \sum_i c^{(l)}_{ij}W^{(l)}_{ij}\cdot x^{(l)}_i \Vert^2}
\end{equation}

In this way, the energy function defined in Equation~\ref{eqn:product-of-capsules-energy} is consistent with routing by agreement, in the sense that the magnitude of the squashed capsule represents the probability that the individual capsule is on. 

\begin{equation}
    P\left( j = \textnormal{on} | x^{(l)}\right) =
    \Vert x^{(l+1)}_j \Vert =
    P\left(\Vert x^{(l+1)}_j \Vert = 1 | x^{(l)}\right)
\end{equation}
where again $x^{(l+1)}_j = \textnormal{squash}\left( z^{(l+1)}_j \right)$ for $z^{(l+1)}_j = \sum_i c^{(l)}_{ij}W^{(l)}_{ij}\cdot x^{(l)}_i $.

Because of this consistency, we use dynamic routing between capsules for mixing in the MCMC estimation of the distribution. Note that we do not want to directly sample $x^{(l+1)}_j$ from a graphical model $P \left( x^{(l+1)}_j | x^{(l)}_i \right)$. This is because $x^{(l+1)}_j$ is a squashed vector, implying that the components of the vector are dependent on each other from the squashing, so graphically the nodes at that given layer are connected, meaning that we cannot efficiently sample these nodes with MCMC. By explicitly decoupling $x^{(l+1)}_j$ into its magnitude and orientation, we are able to decouple the parts of the dynamic routing that we want to use from the parts we do not want to use, allowing us to efficiently sample the magnitude $ \Vert x^{(l+1)}_j \Vert \sim P \left( \Vert x^{(l+1)}_j \Vert | x^{(l)}_i \right)$ from this distribution, since graphically these nodes are not connected, from Equation~\ref{eqn:capsules-not-connected}.

For inference, the energy model only requires $ P\left( \Vert x^{(l+1)}_j \Vert = 1 | x^{(l)}\right) = \Vert x^{(l+1)}_j \Vert $, which holds by construction. We then sample $ x^{(l+1)}_j \sim P\left( x^{(l+1)}_j | x^{(l)}\right) $ as $x^{(l+1)}_j =  \textnormal{squash} \left( \sum_i c^{(l)}_{ij}W^{(l)}_{ij}\cdot x^{(l)}_i \right)$, from the dynamic routing, and again the norm of this is consistent with the energy model. Taking the norm of $x^{(l+1)}_j$ introduces a rotational invariance to the vector, which at first may seem like a problem since the instantiation parameters are stored in the rotational angles of the capsule. In fact this is not an issue, as we will see in Section~\ref{sec:grad-of-log-like}, because we are taking the gradient of the log likelihood, and the gradient of the norm of a vector is dependent on the orientation of the vector itself, not just its magnitude. In this way, during the gradient descent, information from the orientation of the capsule is used to update the model parameters.

% In order to perform Gibbs sampling, we need to be able to sample points from the Markov chain, which itself requires sampling the $x^{(l)}_i$ capsules from the $x^{(l+1)}_j$'s. This is the purpose of the as-of-yet unused $E_2\left( \hat{x}^{(l+1)}, \Vert x^{(l)} \Vert \right)$ term defined in the energy functional. Following the procedure as above, one has in the reverse direction:

% \begin{equation}
% P\left(\Vert x^{(l)}_i \Vert = 1 | \hat{x}^{(l+1)}\right) =
% \frac{\Vert \sum_j c^{(l)}_{ij}W^{(l)T}_{ij}\cdot \hat{x}^{(l+1)}_j \Vert^2 }{1+\Vert \sum_j c^{(l)}_{ij}W^{(l)T}_{ij}\cdot \hat{x}^{(l+1)}_j \Vert^2}
% \end{equation}

\subsection{Gradient of the log-likelihood}
\label{sec:grad-of-log-like}

In order to optimize our parameter weights we need the gradient of the log-likelihood to be tractably computable; the log-likelihood is given by:
\begin{equation}
    \log P\left(  x^{(l)} \right) =
    \log \sum^1_{\Vert x^{(l+1)} \Vert =0} e^{-E\left( x^{(l)} , \Vert x^{(l+1)} \Vert \right)} 
    - 
    \log 
    \sum_{x^{(l)}}
    \sum^1_{\Vert x^{(l+1)} \Vert =0} e^{-E\left( x^{(l)} , \Vert x^{(l+1)} \Vert \right)} 
\end{equation}
We can put the gradient of the log-likelihood in a compact form using the fact that $ P\left( \Vert x^{(l+1)} \Vert | x^{(l)} \right) = 
\nicefrac{P\left( x^{(l)} , \Vert x^{(l+1)} \Vert \right)}{P\left( x^{(l)} \right)} = \\
\nicefrac{\frac{1}{Z} 
\exp{-E\left( x^{(l)} , \Vert x^{(l+1)} \Vert \right)} }
{\frac{1}{Z}
\sum^1_{\Vert x^{(l+1)} \Vert =0} \exp{-E\left( x^{(l)} , \Vert x^{(l+1)} \Vert \right)}
}$.
\begin{multline}
    \frac{\partial}{\partial W^{(l)}_{ij}}
    \log P\left(  x^{(l)} \right) =
    \sum^1_{\Vert x^{(l+1)} \Vert =0} 
    P\left( \Vert x^{(l+1)} \Vert | x^{(l)} \right)
    \frac{\partial E }
    {\partial W^{(l)}_{ij}}
    \left( x^{(l)} , \Vert x^{(l+1)} \Vert \right)
    - \\
    \sum_{x^{(l)}} 
    P\left( x^{(l)} \right)
    \sum^1_{\Vert x^{(l+1)} \Vert =0} 
    P\left( \Vert x^{(l+1)} \Vert | x^{(l)} \right)
    \frac{\partial E }
    {\partial W^{(l)}_{ij}}
    \left( x^{(l)} , \Vert x^{(l+1)} \Vert \right)
\end{multline}
Computing this is still not tractable as it involves an exponential number of sums. However, since the layer $l+1$ capsule activations are not connected, from Equation~\ref{eqn:capsules-not-connected}, we can apply an analogous factorization trick to that which is used in learning Restricted Boltzmann Machines. First we reduce the energy function:
\begin{multline}
    \frac{\partial E }
    {\partial W^{(l)}_{ij}}
    \left( x^{(l)} , \Vert x^{(l+1)} \Vert \right) =
    \frac{\partial}
    {\partial W^{(l)}_{ij}}
    \sum_{j'}
    \log \left( \Vert  
    \sum_{i'} c^{(l)}_{i'j'}W^{(l)}_{i'j'}\cdot x^{(l)}_{i'}
    \Vert^2 \right)
    \Vert x^{(l+1)}_{j'} \Vert
    =
    \\
    \frac{\partial}
    {\partial W^{(l)}_{ij}}
    \log \left( \Vert  
    \sum_{i'} c^{(l)}_{i'j}W^{(l)}_{i'j}\cdot x^{(l)}_{i'}
    \Vert^2 \right)
    \Vert x^{(l+1)}_{j} \Vert
\end{multline}
Using this, we are in a position to find the tractable gradient for the product of expert capsules:
\begin{multline}
\label{eqn:tractable-learning-rule}
    \sum^1_{\Vert x^{(l+1)} \Vert =0} 
    P\left( \Vert x^{(l+1)} \Vert | x^{(l)} \right)
    \frac{\partial E }
    {\partial W^{(l)}_{ij}}
    \left( x^{(l)} , \Vert x^{(l+1)} \Vert \right)
    = \\
    \sum^1_{\Vert x^{(l+1)}_{j} \Vert =0} 
    \hspace{-0.75em}
    P\left( \Vert x^{(l+1)}_j \Vert | x^{(l)} \right)
    \frac{\partial}
    {\partial W^{(l)}_{ij}}
    \log \left( \Vert  
    % \sum_{i'} c^{(l)}_{i'j}W^{(l)}_{i'j}\cdot x^{(l)}_{i'}
    z^{(l+1)}_{j}
    \Vert^2 \right)
    \Vert x^{(l+1)}_{j} \Vert
    \hspace{-0.75em}
    \sum^1_{\Vert x^{(l+1)}_{-j} \Vert =0}
    \hspace{-0.75em}
    P\left( \Vert x^{(l+1)}_{-j} \Vert | x^{(l)} \right)
    \\ = 
    P\left( \Vert x^{(l+1)}_j \Vert = 1 | x^{(l)} \right)
    \frac{\partial}
    {\partial W^{(l)}_{ij}}
    \log \left( \Vert  
    % \sum_{i'} c^{(l)}_{i'j}W^{(l)}_{i'j}\cdot x^{(l)}_{i'}
    z^{(l+1)}_{j}
    \Vert^2 \right)
    \end{multline}
where $ x^{(l+1)}_{-j} $ is analogous notation to the RBM case~\cite{hinton2002training}, and refers to all capsules at layer $l+1$ other than the $j^{th}$ capsule. 

To further reduce Equation~\ref{eqn:tractable-learning-rule}, where we write $z^{(l+1)}_{j} = \sum_{i'} c^{(l)}_{i'j}W^{(l)}_{i'j}\cdot x^{(l)}_{i'}$, one has $ \frac{\partial}
{\partial W^{(l)}_{ij}}
\log \left( \Vert z^{(l+1)}_{j} \Vert^2 \right) = \frac{2 c^{(l)}_{ij}}{\Vert  
z^{(l+1)}_{j} \Vert^2} z^{(l+1)}_{j} x^{(l)}_i $. Similarly, $ P\left( \Vert x^{(l+1)}_j \Vert = 1 | x^{(l)} \right) = \frac{\Vert z^{(l+1)}_{j} \Vert^2}{1+\Vert z^{(l+1)}_{j} \Vert^2} $, and we have the final form of the gradient:

\begin{equation}
\label{eqn:gradient_update}
    \frac{\partial}{\partial W^{(l)}_{ij}}
    \log P\left( x^{(l)} \right) =
    2 c^{(l)}_{ij}
    \left(
    \frac{z^{(l+1)}_{j} x^{(l)}_i}{1+\Vert  
    z^{(l+1)}_{j} \Vert^2}
    -
    \sum_{x^{(l)}}
    P\left( x^{(l)} \right)
    \frac{z^{(l+1)}_{j} x^{(l)}_i}{1+\Vert  
    z^{(l+1)}_{j} \Vert^2}
    \right)
    % P\left( \Vert x^{(l+1)}_j \Vert = 1 | \hat{x}^{(l)} \right)
    % \frac{\partial}
    % {\partial W^{(l)}_{ij}}
    % \log \left( \Vert  
    % \sum_{i'} c^{(l)}_{i'j}W^{(l)}_{i'j}\cdot \hat{x}^{(l)}_{i'}
    % \Vert^2  \right) - \\
    % \sum_{\hat{x}^{(l)}}
    % \log P\left(  \hat{x}^{(l)} \right)
    % P\left( \Vert x^{(l+1)}_j \Vert = 1 | \hat{x}^{(l)} \right)
    % \frac{\partial}
    % {\partial W^{(l)}_{ij}}
    % \log \left( \Vert  
    % \sum_{i'} c^{(l)}_{i'j}W^{(l)}_{i'j}\cdot \hat{x}^{(l)}_{i'}
    % \Vert^2 \right)
\end{equation}

For RBM learning, this is analogous to the update rule $ \Delta w_{ij} = \left\langle v_i h_j \right\rangle_{\textnormal{data}} - \left\langle v_i h_j \right\rangle_{\textnormal{model}} $. 

Thus, beginning with the data $x^{(l)}_i$, we use dynamic routing to sample $z^{(l+1)}_{j} = \sum_{i} c^{(l)}_{ij}W^{(l)}_{ij}\cdot x^{(l)}_{i}$ where the $c^{(l)}_{ij}$'s are determined by routing by agreement and $\sum_i c^{(l)}_{ij} = 1$. 

We then squash this $x^{(l+1)}_{j} = \textnormal{squash}\left( z^{(l+1)}_{j} \right)$ and run dynamic routing in reverse to produce $\tilde{z}^{(l)}_{i} = \sum_{j} c^{(l)}_{ij}W^{(l)T}_{ij}\cdot x^{(l+1)}_{j}$, using the same $c^{(l)}_{ij}$'s as before, with $\sum_i c^{(l)}_{ij} = 1$. This is squashed producing $ \tilde{x}^{(l)}_{i}$.
Finally, we repeat the first step, again with the same $c^{(l)}_{ij}$'s such that $\sum_i c^{(l)}_{ij} = 1$ to produce $\tilde{z}^{(l+1)}_{j} = \sum_{i} c^{(l)}_{ij}W^{(l)}_{ij}\cdot \tilde{x}^{(l)}_{i}$ and its squashed counterparts $ \tilde{x}^{(l+1)}_{j} $. This procedure is summarized in Algorithm~\ref{alg:algorithm1}.

We then use these values, which we are referring to as mixing by dynamic routing, for the gradient in Equation~\ref{eqn:gradient_update}. As is usual, instead of minimizing the log likelihood we minimize the contrastive divergence, and so we only run this mixing once. The complete process is summarized in Algorithm~\ref{alg:algorithm1}.

\begin{algorithm}[]
\SetAlgoLined
\KwResult{Trained model parameters $W^{(l)}_{ij}$.}
 Given: $x^{(l)}_i$ \\
 Randomly initialize (e.g.) $W^{(l)}_{ij}\sim N\left(0,0.01\right)$ \\
 \For{\textnormal{number epochs}}{
  $ c^{(l)}_{ij} = \textnormal{RoutingByAgreement}\left(x^{(l)}_i, W^{(l)}_{ij} \right) $ \# find routing coefficients  \\
  $z^{(l+1)}_{j} = \sum_{i} c^{(l)}_{ij}W^{(l)}_{ij}\cdot x^{(l)}_{i}$ \# route capsules forward  \\
  $x^{(l+1)}_{j} = \textnormal{squash}\left( z^{(l+1)}_{j} \right)$  \# squash
  \\
  $\tilde{z}^{(l)}_{i} = \sum_{j} c^{(l)}_{ij}W^{(l)T}_{ij}\cdot x^{(l+1)}_{j}$ \# reconstruct input capsules  \\
  $\tilde{x}^{(l)}_{i} = \textnormal{squash}\left( \tilde{z}^{(l)}_{i} \right)$ \# squash 
  \\
  $\tilde{z}^{(l+1)}_{j} = \sum_{i} c^{(l)}_{ij}W^{(l)}_{ij}\cdot \tilde{x}^{(l)}_{i}$ \# reconstruct output capsules \\
  $\tilde{x}^{(l+1)}_{j} = \textnormal{squash}\left( \tilde{z}^{(l+1)}_{j} \right)$ \# squash 
  \\
  $\Delta W^{(l)}_{ij} =
    2 c^{(l)}_{ij}
    \left(
    \frac{z^{(l+1)}_{j} x^{(l)}_i}{1+\Vert  
    z^{(l+1)}_{j} \Vert^2}
    -
    \frac{\tilde{z}^{(l+1)}_{j} \tilde{x}^{(l)}_i}{1+\Vert  
    \tilde{z}^{(l+1)}_{j} \Vert^2}
    \right)$ \# calculate gradient \\
    $ W^{(l)}_{ij} \xleftarrow[]{} W^{(l)}_{ij} + \lambda \Delta W^{(l)}_{ij}$ \# update weights \
 }
 \caption{Training products of expert capsules. In practice we use learning rate decay, sgd with momentum and an $\ell_2$-regularization on the weights for the weight update.}
 \label{alg:algorithm1}
\end{algorithm}

\section{Experiments}

\subsection{Network architecture}

\begin{figure}[t]
    \centering
    \includegraphics[width=0.93\textwidth]{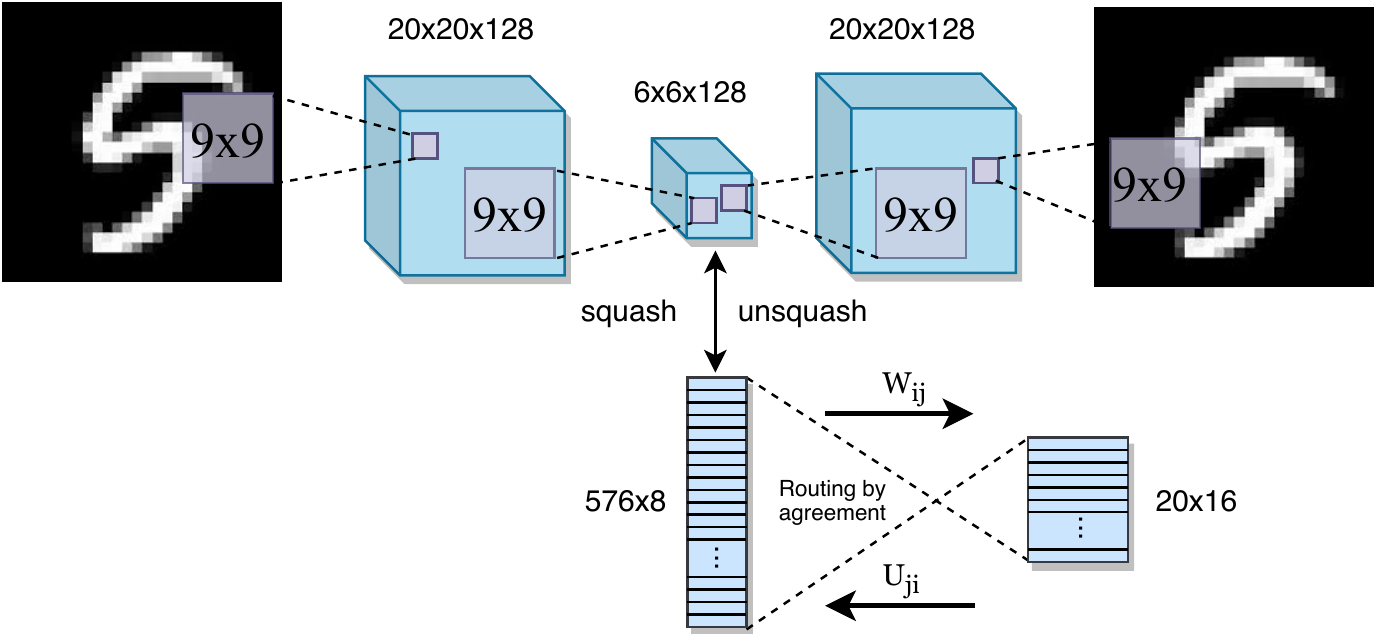}
    \caption{The network architecture used for these experiments. First a convolutional autoencoder is trained to learn the filters. Next, with the filter weights fixed, we proceed to learn the capsule encoder weights $W_{ij}$. Once the capsule encoder is trained, we learn the capsule decoder weights $U_{ji}$. To generate the images, we randomly sample points from the $16$-dimensional encoded capsule space, squash, dynamically route with the $U_{ji}$'s to the $8$-dimensional decoded capsule space, unsquash and reshape, and finally use the deconvolutional decoder to generate the images.}
    \label{fig:network-architecture}
\end{figure}

The outline of the experimental architecture can be seen in Figure~\ref{fig:network-architecture}. Other than the input/output channels being greyscale and rgb, the same architecture was used for both experiments. First an autoencoding convolutional network~\cite{zeiler2014visualizing}, with dropout~\cite{srivastava2014dropout} was used to learn the convolutional filter weights in an unsupervised way. The first filter bank is of size $\left[9,9,1,128\right]$ (for height $\times$ width $\times$ channels in $\times$ channels out), followed by a leaky ReLU activation, while the second filter bank is $\left[9,9,128,128\right]$ followed by a leaky ReLU activation. The transposed filters are used in the decoder, with first a leaky ReLU and a sigmoid at the output, to scale the pixel intensities between $0$ and $1$.

Once the convolutional autoencoder is sufficiently trained, these weights are held fixed and the $128$-dimensional hidden layer is reshaped to $6\times 6\times 128/8=576$ capsules, each being $8$-dimensional, and then squashed along these $8$ dimensions. Using the unsupervised product of expert capsules training algorithm described above, these capsules are mapped to $20$ capsules, each of $16$-dimensions, to learn the $W^{(l)}_{ij}$'s in $z^{(l+1)}_j=\sum_i c^{(l)}_{ij} W^{(l)}_{ij} \cdot x^{(l)}_{i}$, where $i=1,2,\dots,576$ and $j=1,2,\dots,20$.

After the $W^{(l)}_{ij}$'s are trained, these weights are held fixed and we learn a decoder $z^{(l)}_i=\sum_j e^{(l+1)}_{ji} U^{(l+1)}_{ji} \cdot x^{(l+1)}_{j}$ so that we can sample from the $16$-dimensional capsule space to generate the images. The update rule is $ \Delta U^{(l+1)}_{ji} = 2e^{(l+1)}_{ji} \left( \frac{z^{(l)}_{i}x^{(l+1)}_{j}}{1+\Vert z^{(l)}_{i} \Vert^2} - 
\frac{\tilde{z}^{(l)}_{i}\tilde{x}^{(l+1)}_{j}}{1+\Vert \tilde{z}^{(l)}_{i} \Vert^2}
\right) $, where (the data) $z^{(l)}_{i}$ and $x^{(l+1)}_{j}$ are fixed by the $W^{(l)}_{ij}$'s, while the (the model) $\tilde{z}^{(l)}_{i}=\sum_j e^{(l+1)}_{ji} U^{(l+1)}_{ji} x^{(l+1)}_{j} $ and $\tilde{x}^{(l+1)}_{j} = \textnormal{squash}\left( \sum_j e^{(l+1)}_{ji} U^{(l+1)T}_{ji} \tilde{x}^{(l)}_{j} \right)$ are generated with the $U^{(l+1)}_{ji}$'s. We implemented this network in TensorFlow~\cite{abadi2016tensorflow}, and training on a single gpu took about fifteen minutes.

\subsection{Routing-Weighted Product of Expert Neurons}

\begin{figure}
     \centering
     \begin{subfigure}[b]{1.\textwidth}
        \includegraphics[width=\textwidth]{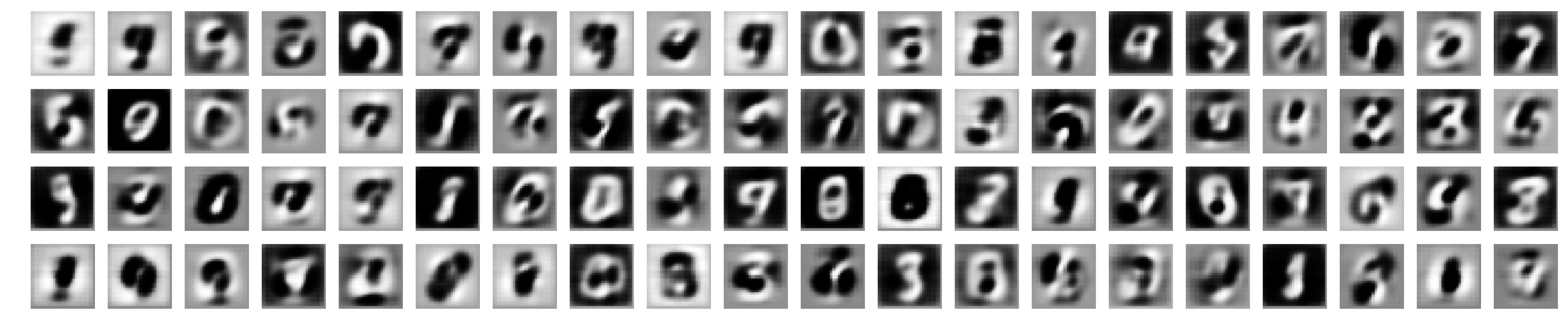}
         \caption{MNIST - sampled from complete domain.}
         \label{fig:2a}
     \end{subfigure}
     \hfill
     \begin{subfigure}[b]{1.\textwidth}
         \centering
        \includegraphics[width=\textwidth]{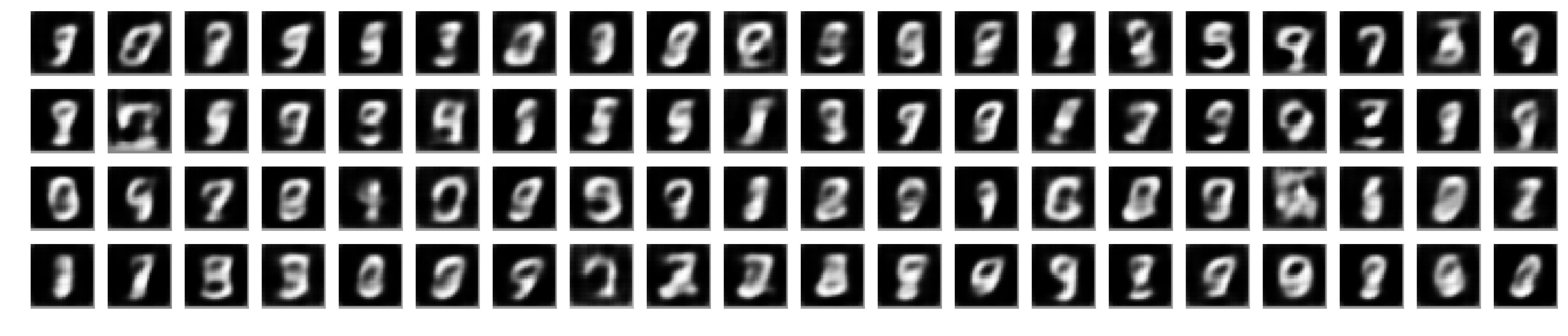}
         \caption{MNIST - sampled from restricted domain.}
         \label{fig:2b}
     \end{subfigure}
     \hfill
     \begin{subfigure}[]{1.\textwidth}
         \centering
        \includegraphics[width=\textwidth]{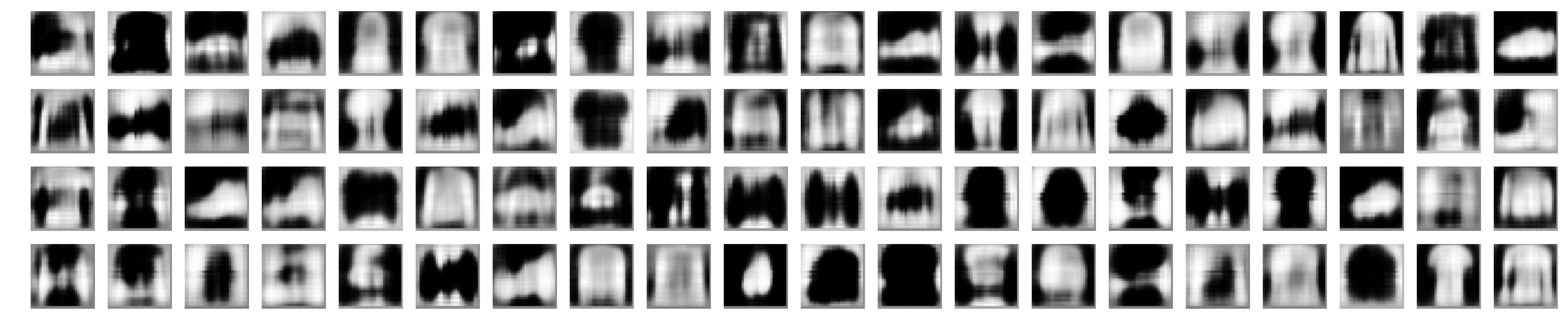}
         \caption{Fashion-MNIST - sampled from complete domain.}
         \label{fig:2c}
     \end{subfigure}
     \hfill
     \begin{subfigure}[]{1.\textwidth}
         \centering
        \includegraphics[width=\textwidth]{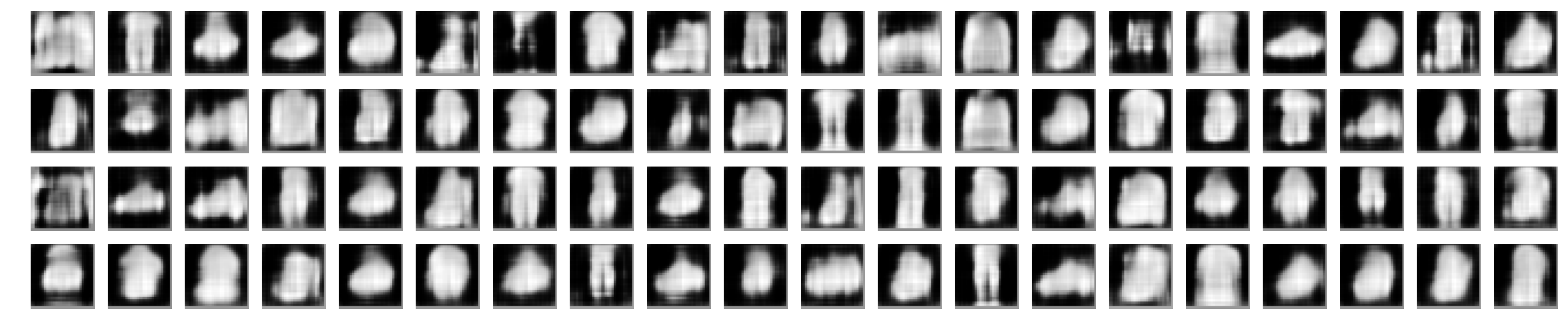}
         \caption{Fashion-MNIST - sampled from restricted domain.}
         \label{fig:2d}
     \end{subfigure}
    \caption{Images created by the unsupervised, routing-weighted product of expert neurons model. For each model, all of the 80 images were sampled together so these are not cherry picked examples. Each column is one of the twenty hidden layer capsules, while the four rows are four random samplings for that $16$-dimensional capsule. It is seen that individual capsules learn specific objects in this unsupervised setting, and different samples drawn from these capsules, i.e. different instantiation parameters, yield changes in the object. For example in MNIST these instantiation dimensions yield changes in stroke thickness, angles and lengths, while in Fashion-MNIST they can transform dimensions such as sleeve length and height of the shoes.}
    \label{fig:routing-weighted-poec}
\end{figure}

In Figure~\ref{fig:routing-weighted-poec} there are $20$ columns for each of the $20$ capsules, and the $4$ rows are random samples from a $16$-dimensional Gaussian distrbition, with the other $19$ capsules having $0$ as their input. %A positive uniform distribution is used, as opposed to say a Gaussian, because we wanted to restrict the points to a smaller region of the entire $16$-dimensional space to help understand the subtle differences in the dimensions that the capsules are representing; instead of allowing all possible angular instantiation parameters, we restrict it to those between $0$ and $90$ degrees.
It is seen that each of the individual capsules learn specific objects, with different random samples generating images with different instantiation parameters of similar objects, such as stroke thicknesses and angles, or sleeves coming out of pants to create shirts and jackets. Interestingly some of the images are negatives of what is expected, where if the input $x\sim N\left(0,1\right)$ is replaced with $-x$ the image generated are no longer negatives, but the ones that were normal become negatives, as is seen in Figures~\ref{fig:2a} and~\ref{fig:2c}.

We believe this stems from the fact that the $16$ dimensional point, as understood by the capsule, lives on the manifold $S^{15}\times\left(0,1\right)$, where $S^{15}$ is the $15$-dimensional sphere of angular orientations, and the $\left(0,1\right)$ is the capsule magnitude representing off or on. Not all object instantiation parameters should live on the sphere. For example rotating an object in a circle should live on $S^1$, but increasing the size of an object should live on $\mathbb{R}$, since an object's size shouldn't return to where it started if the size is monotonically increasing, as would happen on $S^1$. In this way the pixel intensities can be reversed $180^\circ$ since they are coming from $S^{15}$ as opposed to $\mathbb{R}^{15}$. When we restrict our sampling domain to the half of $S^{15}$ that is visited during training the problem is alleviated, as seen in Figures~\ref{fig:2b} and~\ref{fig:2d}, suggesting that this idea is infact correct.

A more elegant solution, although outside the scope of this paper, would be to design a capsule on, say, $S^m \times \mathbb{R}^n \times \left(0,1\right)$. Nevertheless, the unsupervised learning procedure developed here for the capsules is distinctly learning recognizable objects.

\section{Conclusions}

This work developed capsule networks with routing by agreement within a product of experts formulation. Observing that the magnitudes of the hidden layer capsules are not connected, we design an energy function that is consistent with dynamic routing, in the sense that the binary action of a hidden layer capsule firing is equal to the probability of a capsule being on, when calculated with dynamic routing by agreement. We then use dynamic routing to mix the distribution. The gradient of the log likelihood is found and used to minimize the contrastive divergence. A simple network architecture is set up to test this unsupervised learning algorithm, and is able to generate images similar to those of the datasets it was trained on.

\newpage

% \bibliographystyle{unsrt}
% \bibliography{bibliography}

\end{document}